\documentclass[conference]{IEEEtran}
\IEEEoverridecommandlockouts
\usepackage{cite}
\usepackage{amsmath,amssymb,amsfonts}
\usepackage{algorithmic}
\usepackage{graphicx}
\usepackage{textcomp}
\usepackage{xcolor}

\usepackage{booktabs}
\usepackage{multirow}

\usepackage[caption=false,font=footnotesize]{subfig}

\usepackage{pgfplots}
\pgfplotsset{compat=1.18}
\usepackage{tikz}
\usetikzlibrary{shapes, arrows, positioning, fit, calc, backgrounds}
\def\BibTeX{{\rm B\kern-.05em{\sc i\kern-.025em b}\kern-.08em
    T\kern-.1667em\lower.7ex\hbox{E}\kern-.125emX}}

\renewcommand{\thetable}{\arabic{table}}
\renewcommand{\thefigure}{\arabic{figure}}
\makeatletter
\def\fnum@figure{Figure~\thefigure}
\def\fnum@table{Table~\thetable}
\makeatother

\let\origcite\cite
\renewcommand{\cite}[1]{\textsuperscript{\origcite{#1}}}

\begin{document}

\title{Non-Intrusive Graph-Based Bot Detection for E-Commerce Using Inductive Graph Neural Networks}

\author{\IEEEauthorblockN{Sichen Zhao\textsuperscript{*}}
\IEEEauthorblockA{\textit{College of Engineering}\\
\textit{Northeastern University}\\
Boston, USA\\
zhao.siche@northeastern.edu}
\and
\IEEEauthorblockN{Zhiming Xue}
\IEEEauthorblockA{\textit{College of Engineering}\\
\textit{Northeastern University}\\
Boston, USA\\
xue.zh@northeastern.edu}
\and
\IEEEauthorblockN{Yalun Qi}
\IEEEauthorblockA{\textit{Khoury College of Computer Sciences}\\
\textit{Northeastern University}\\
Boston, USA\\
qi.yal@northeastern.edu}
\and
\IEEEauthorblockN{Xianling Zeng}
\IEEEauthorblockA{\textit{College of Engineering}\\
\textit{Northeastern University}\\
Boston, USA\\
zeng.xian@northeastern.edu}
\and
\IEEEauthorblockN{Zihan Yu}
\IEEEauthorblockA{\textit{College of Professional Studies}\\
\textit{Northeastern University}\\
Boston, USA\\
yu.zihan1@northeastern.edu}
}

\maketitle

\begin{abstract}
Malicious bots abuse e-commerce services while evading conventional defenses. IP/rule blocking is brittle under proxy rotation, and CAPTCHAs add friction yet are often bypassed. We propose a non-intrusive framework that models session--URL interactions as a bipartite graph and uses an inductive GNN (GraphSAGE) to classify session nodes. Combining topology with lightweight behavioral and URL semantics enables detection of ``feature-normal'' automation. On real-world traffic with high-confidence bot labels, GraphSAGE outperforms a session-feature MLP baseline in AUC and F1, and remains robust under mild adversarial edge perturbations and in cold-start inductive evaluation---supporting real-time deployment without client-side instrumentation.
\end{abstract}

\begin{IEEEkeywords}
bot detection, graph neural networks, e-commerce security, GraphSAGE, fraud detection, machine learning
\end{IEEEkeywords}

\section{Introduction}
E-commerce platforms face persistent automated abuse from bots that mimic human browsing. Common defenses are brittle under proxy rotation or intrusive (CAPTCHAs), motivating passive detection from backend telemetry.

Graphs compactly capture relationships across sessions and content, enabling detection beyond per-session aggregates (e.g., BotChase \cite{aboudaya2020botchase}). We construct a bipartite session--URL interaction graph from standard logs and apply an inductive GNN, GraphSAGE \cite{hamilton2017inductive}, to classify session nodes. We evaluate accuracy gains and robustness to graph perturbations and temporal shift, including cold-start scoring for unseen sessions and pages.

Our main contributions are summarized as follows:

\begin{itemize}
\item \textbf{Non-intrusive graph formulation:} We formulate bot detection on a session--URL graph from standard server logs, avoiding CAPTCHAs and client-side instrumentation.

\item \textbf{Inductive GraphSAGE with lightweight features:} We use GraphSAGE over session and URL attributes to score unseen session/URL nodes.

\item \textbf{Robustness + deployability evaluation:} We report gains over a session-feature MLP baseline and study adversarial edge perturbations and cold-start generalization.
\end{itemize}

The remainder of this paper reviews related work (Section~\ref{sec:related}), presents the method (Section~\ref{sec:method}), reports experiments including perturbation and cold-start simulations (Section~\ref{sec:experiments}), analyzes results and deployment considerations (Section~\ref{sec:results}), and concludes (Section~\ref{sec:conclusion}).

\section{Related Work}
\label{sec:related}

\subsection{Bot Detection in Web and E-commerce}
Traditional web bot defenses rely on static indicators or challenges, but modern bots evade them via proxies and automation. Recent work thus emphasizes behavior-driven ML detection to improve adaptability while reducing friction \cite{suchacka2021webbot,iliou2021webbots}; we follow this passive, high-precision direction.

\subsection{Graph-Based Fraud and Bot Detection}
Relational structure is often diagnostic in security and fraud. BotChase \cite{aboudaya2020botchase} shows improved robustness via graph learning, and GNNs are increasingly used in fraud workflows where relational context improves detection and operational utility \cite{huo2025fraudgnn}, motivating session--content graphs for e-commerce.

\subsection{Graph Neural Networks in Anomaly Detection}
GNNs are widely used for graph anomaly detection \cite{ma2021comprehensive}, but production graphs are dynamic and require inductive handling of unseen nodes. GraphSAGE \cite{hamilton2017inductive} learns feature-driven aggregators that generalize beyond the training graph, fitting live e-commerce traffic.

\section{Method}
\label{sec:method}

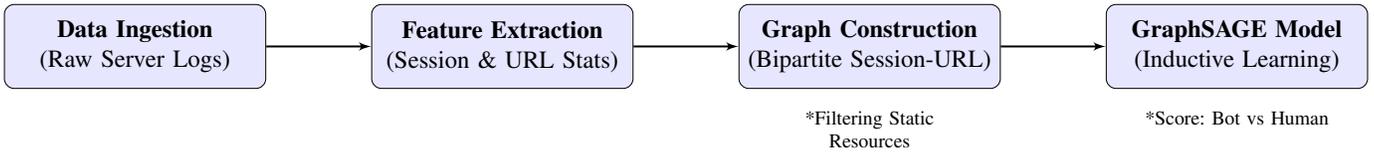
\begin{figure*}[htbp] 
    \centering
    \resizebox{\textwidth}{!}{%
    \begin{tikzpicture}[
        node distance=1.5cm,
        auto,
        block/.style={rectangle, draw, fill=blue!10, text width=3.5cm, text centered, rounded corners, minimum height=1.2cm},
        line/.style={draw, -latex', thick}
    ]
        \node [block] (logs) {\textbf{Data Ingestion}\\ (Raw Server Logs)};
        \node [block, right=of logs] (feature) {\textbf{Feature Extraction}\\ (Session \& URL Stats)};
        \node [block, right=of feature] (graph) {\textbf{Graph Construction}\\ (Bipartite Session-URL)};
        \node [block, right=of graph] (model) {\textbf{GraphSAGE Model}\\ (Inductive Learning)};
        
        \path [line] (logs) -- (feature);
        \path [line] (feature) -- (graph);
        \path [line] (graph) -- (model);
        
        \node [below=0.2cm of graph, text width=3cm, text centered, font=\footnotesize] {*Filtering Static Resources};
        \node [below=0.2cm of model, text width=3cm, text centered, font=\footnotesize] {*Score: Bot vs Human};
        
    \end{tikzpicture}%
    }
    \caption{\textbf{Overall Architecture.} The proposed framework transforms raw logs into behavioral features, constructs a filtered bipartite interaction graph, and utilizes an inductive GraphSAGE model for real-time bot detection.}
    \label{fig:architecture}
\end{figure*}

Our method comprises graph construction, feature design, model architecture, and training/inference. Figure~\ref{fig:architecture} summarizes the end-to-end pipeline.

\subsection{Graph Construction}
We construct a heterogeneous bipartite graph $G=(V,E)$ that links sessions to accessed URLs.

\begin{itemize}
\item \textbf{Session nodes:} Each node is a user session (a sequence of requests/actions within a time window).

\item \textbf{Content/URL nodes:} Each node is a unique page/resource (e.g., product, category, search).
\end{itemize}

An edge $e\in E$ connects a session node to a URL node if the session accessed the URL. For message passing we treat edges as undirected and unweighted; repeated requests are captured by session features rather than edge multiplicity.

\begin{figure}[htbp]
    \centering
    \resizebox{\columnwidth}{!}{%
    \begin{tikzpicture}[scale=0.8]
        \foreach \i in {1,...,4} {
            \node[circle, draw, fill=blue!20, minimum size=0.6cm] (s\i) at (0, -\i*1.2) {$S_\i$};
        }
        \node[above=0.2cm] at (0, -0.8) {\textbf{Sessions}};
        
        \foreach \j in {1,...,3} {
            \node[rectangle, draw, fill=orange!20, minimum size=0.6cm] (u\j) at (4, -\j*1.5 - 0.5) {$U_\j$};
        }
        \node[above=0.2cm] at (4, -1.5) {\textbf{URLs}};
        
        \draw (s1) -- (u1);
        \draw (s1) -- (u2);
        \draw (s2) -- (u2); 
        \draw (s3) -- (u3);
        \draw (s4) -- (u3); 
        \draw[dashed, red] (s4) -- (u1); 
        
    \end{tikzpicture}%
    }
    \caption{Illustration of the bipartite session--URL interaction graph used in this work. Nodes represent sessions and accessed URLs, and edges indicate page visits.}
    \label{fig:graph_structure}
\end{figure}
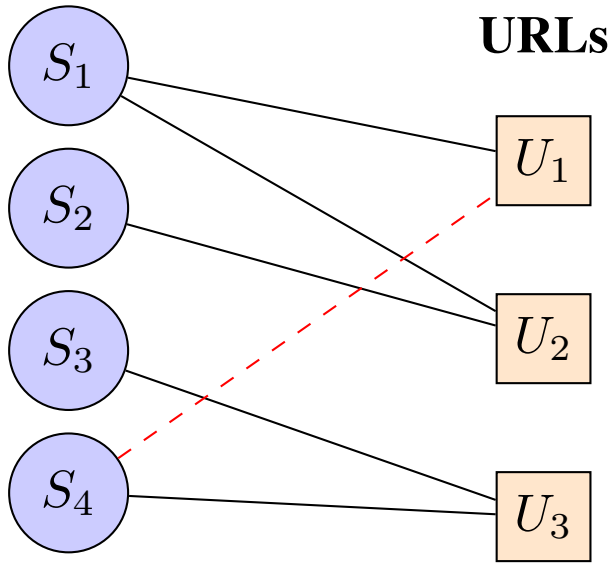

Figure~\ref{fig:graph_structure} illustrates the bipartite session--URL graph used throughout this work.

\textbf{Motivation:} Legitimate sessions follow common navigation patterns, while bots often induce atypical connectivity (broad coverage, rare-page combinations, or coordinated targeting). The graph supports ``suspicion by association'' via shared URL neighborhoods \cite{huo2025fraudgnn}.

We parse logs, map requests to sessions, add nodes for unique sessions/URLs, and update edges online.

\begin{figure}[htbp]
    \centering
    \subfloat[Raw Graph (Noisy Hubs)]{%
        \resizebox{0.45\columnwidth}{!}{%
        \begin{tikzpicture}[scale=0.55, every node/.style={scale=0.6}]
            \node[circle, fill=red!30, draw] (hub) at (0,0) {CSS};
            \foreach \angle in {0,45,...,315} {
                \node[circle, fill=blue!20, draw, inner sep=1pt] (n\angle) at (\angle:1.5) {};
                \draw (hub) -- (n\angle);
            }
        \end{tikzpicture}%
        }
    }\hfill
    \subfloat[Refined Graph (Clean)]{%
        \resizebox{0.45\columnwidth}{!}{%
        \begin{tikzpicture}[scale=0.55, every node/.style={scale=0.6}]
            \node[circle, fill=orange!30, draw] (u1) at (-0.5,0.5) {P1};
            \node[circle, fill=orange!30, draw] (u2) at (0.5,-0.5) {P2};

            \node[circle, fill=blue!20, draw, inner sep=1pt] at (-1.2, 0.8) {} edge (u1);
            \node[circle, fill=blue!20, draw, inner sep=1pt] at (-0.8, 1.2) {} edge (u1);

            \node[circle, fill=blue!20, draw, inner sep=1pt] at (1.2, -0.8) {} edge (u2);
            \node[circle, fill=blue!20, draw, inner sep=1pt] at (0.8, -1.2) {} edge (u2);

            \draw (u1) -- (u2);
        \end{tikzpicture}%
        }
    }
    \caption{Graph refinement process. (a) Raw graphs are dominated by static resource hubs (e.g., CSS). (b) Filtering yields meaningful clusters.}
    \label{fig:refinement}
\end{figure}
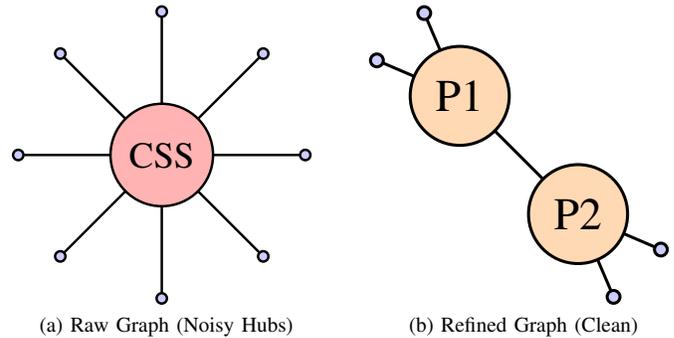

Figure~\ref{fig:refinement} shows why we filter static-resource hubs to obtain a cleaner graph for message passing.

\subsection{Feature Design}
We assign lightweight feature vectors to session and URL nodes.

\begin{figure*}[t]
    \centering
    \includegraphics[width=0.9\textwidth]{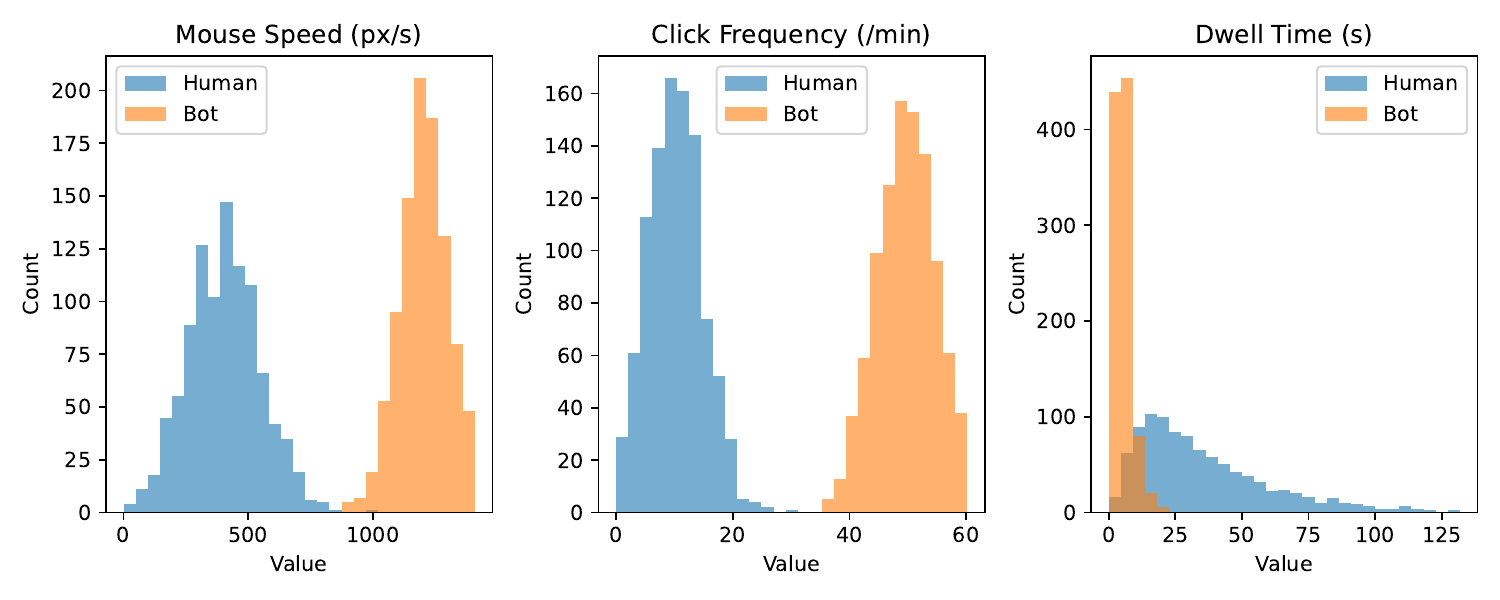}
    \caption{Distribution of representative session-level behavioral features for human and bot sessions. 
    Although individual features exhibit partial overlap, the distributions reveal systematic differences, motivating the use of relational graph modeling.}
    \label{fig:session_features}
\end{figure*}
Figure~\ref{fig:session_features} visualizes representative feature distributions and motivates combining attributes with relational structure.


\textbf{Session node features:}

\begin{itemize}
\item \textit{Temporal and volume signals:} session duration, request count, and request rate.
\item \textit{Coverage and depth:} distinct pages/categories and indicators of multi-step actions (e.g., cart/login).
\item \textit{Lightweight fingerprints:} coarse user-agent/headers when available; we avoid intrusive client-side telemetry.
\end{itemize}

Features are numeric/categorical (standardized/encoded) and kept lightweight for non-intrusive deployment.

\textbf{Content (URL) node features:} Content (URL) node features are intentionally designed to be coarse-grained and privacy-preserving.
We do not use raw URL strings, path tokens, query parameters, or any user-generated content (e.g., search terms or identifiers) as model inputs.
Instead, each URL is mapped to a small set of high-level semantic attributes, including page category (e.g., product, category, search, checkout)
and global access statistics such as relative popularity or rarity.

All URL identifiers are anonymized via one-way hashing prior to graph construction, and the model operates exclusively on these abstracted features.
This design follows data minimization principles and avoids exposure to personally identifiable information (PII), enabling non-intrusive deployment
without client-side instrumentation or explicit user consent.

\begin{itemize}
\item \textit{Type/context:} page category (e.g., product, category, search, checkout).
\item \textit{Global statistics:} relative popularity/rarity and coarse sensitivity tags for special endpoints.
\end{itemize}

These URL attributes contextualize access patterns (e.g., concentrated browsing of rare or sensitive endpoints) while preserving privacy.

\textbf{Motivation:} Session-only models miss ``feature-normal'' bots; message passing combines behavior, page context, and shared neighborhoods.

\subsection{Model Architecture}

We use GraphSAGE \cite{hamilton2017inductive} to learn node representations via sampled neighbor aggregation. Its inductive formulation is critical because new sessions and URLs appear continuously.

\textbf{GraphSAGE layers:} Two layers capture 1-hop and 2-hop context. Each layer updates node $v$ as:
\begin{equation}
\mathbf{h}_v^{(k)} = \sigma\left(W^{(k)} \cdot \text{AGG}\left(\mathbf{h}_v^{(k-1)}, \mathbf{h}_{u_1}^{(k-1)}, \mathbf{h}_{u_2}^{(k-1)}, \ldots\right)\right),
\end{equation}
where $\mathbf{h}_v^{(0)}$ is the input feature vector, neighbors are $\{u_1, u_2, \ldots\}$, $\text{AGG}(\cdot)$ is permutation-invariant, $W^{(k)}$ are trainable weights, and $\sigma$ is a non-linearity (ReLU). We use a mean aggregator.

After two layers we obtain embeddings (e.g., 128-D) that summarize multi-hop neighborhoods: a session representation reflects its own features and the pages it visited, as well as other sessions that visited those pages.

\textbf{Classifier:} We apply an MLP head to session embeddings to output $P(\text{bot} \mid \text{session})$. Only session nodes are supervised; URL nodes participate in message passing as contextual carriers.

\textbf{Inductiveness:} GraphSAGE learns an aggregation function (not per-node embeddings), enabling scoring of new sessions/URLs from features and neighborhoods.

\textbf{Why GraphSAGE:} Neighbor sampling improves scalability on sparse, high-degree graphs, and the inductive formulation supports evolving graphs better than transductive alternatives.

\subsection{Training and Inference}

\textbf{Training:} We train supervised on labeled sessions. Labels are obtained via a hybrid (semi-synthetic) strategy combining verified real-world attacks (e.g., honeypots/trap URLs) with controlled injections of diverse bot scripts; imbalance is handled via weighting/resampling.

We use binary cross-entropy on labeled session nodes:
\begin{equation}
L = -\frac{1}{N}\sum_{i=1}^N \left[y_i \log \hat{y}_i + (1-y_i)\log(1-\hat{y}_i)\right],
\end{equation}
where $y_i$ is the true label and $\hat{y}_i$ the predicted probability. We train with mini-batches by sampling target sessions and their $k$-hop neighborhoods (GraphSAGE sampling), with dropout and $L_2$ weight decay for regularization.

We select hyperparameters on validation AUC and recall under low false positive rates, reflecting operational constraints.

\textbf{Inference:} For a new session, we add its node/edges, compute features, aggregate a bounded neighborhood, and output a bot probability.

Inference runs on a bounded neighborhood subgraph, enabling near real-time scoring.

\textbf{Incremental learning:} The model can be periodically retrained with new confirmed samples and hot-swapped without changing online graph construction.

\section{Experiments}
\label{sec:experiments}

We evaluate on real traffic logs and two simulations to test accuracy gains over a session-feature baseline and robustness to perturbations and unseen nodes.

\subsection{Dataset and Experimental Setup}

\textbf{Dataset:} We use anonymized server logs from a representative mid-sized e-commerce platform over two weeks, built via a hybrid (semi-synthetic) strategy for high-confidence labels. Background traffic consists of real production sessions ($\sim$80K) after removing sessions with $<2$ requests and truncating extreme outliers. Bots constitute a small fraction ($\sim$5\%) and are drawn from verified real-world attacks (honeypots/trap URLs) as well as controlled injections (scrapers/headless browsers). The session--URL graph has on the order of $10^5$ edges (tens of thousands of sessions; thousands of URLs) with a power-law degree distribution. All session and URL identifiers are anonymized using one-way hashing, and no raw URLs, query parameters, or user-specific identifiers are retained in the dataset. We use 10\% validation and 10\% test splits with similar class proportions, and a chronologically later test split to emulate deployment and evaluate inductive generalization.

\textbf{Baselines:} We compare against a session-feature MLP trained on the same session features but without graph connectivity. This isolates the value added by relational modeling.

We report AUC plus precision/recall/F1 at an operating threshold chosen to yield approximately 1\% false positive rate on validation.

\textbf{Training details:} GraphSAGE uses two layers with 128-dimensional hidden states and neighbor sampling size 15; the MLP has two 128-unit hidden layers. Both use Adam (lr=0.001), early stopping on validation AUC, and class weights for imbalance. We run 5 seeds and report mean performance.

\subsection{Performance Comparison}

Table~\ref{tab:performance_comparison} shows that GraphSAGE (refined graph) outperforms the session-only MLP (AUC 0.9705 vs. 0.9102) and improves recall at $\sim$1\% FPR. The raw-graph variant underperforms and is less stable, motivating refinement.

\begin{table}[htbp]
    \centering
    \caption{Overall Bot Detection Performance Comparison on Test Set.}
    \label{tab:performance_comparison}
    \renewcommand{\arraystretch}{1.15} 
    \small
    \resizebox{\columnwidth}{!}{%
    \begin{tabular}{lcccc}
        \toprule
        \textbf{Model} & \textbf{AUC (Mean $\pm$ Std)} & \textbf{Precision} & \textbf{Recall (@1\% FPR)} & \textbf{F1-Score} \\
        \midrule
        Session-level MLP (Baseline) & 0.9102 $\pm$ 0.0150 & 0.7505 & 0.7510 & 0.7508 \\
        GraphSAGE (Raw Graph) & 0.8756 $\pm$ 0.1042 & 0.8230 & 0.8105 & 0.8167 \\
        \textbf{GraphSAGE (Ours, Refined)} & \textbf{0.9705 $\pm$ 0.0085} & \textbf{0.8055} & \textbf{0.9002} & \textbf{0.8501} \\
        \bottomrule
    \end{tabular}%
    }
\end{table}

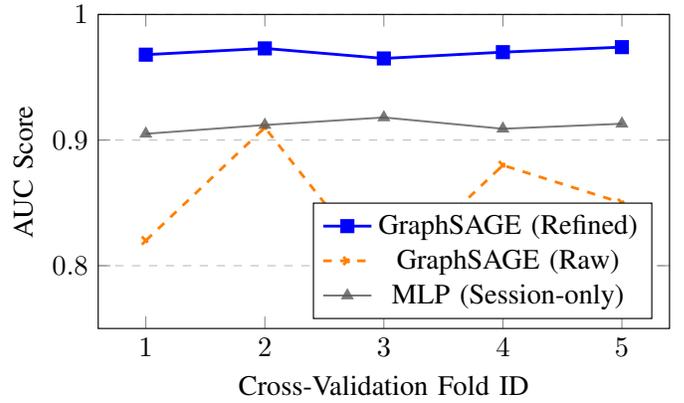
\begin{figure}[htbp]
    \centering
    \resizebox{\columnwidth}{!}{%
    \begin{tikzpicture}
        \begin{axis}[
            width=0.48\textwidth,
            height=5.5cm,
            xlabel={Cross-Validation Fold ID},
            ylabel={AUC Score},
            ymin=0.75, ymax=1.0,
            xtick={1,2,3,4,5},
            legend pos=south east,
            ymajorgrids=true,
            grid style=dashed
        ]
        
        \addplot[color=blue, mark=square*, line width=1pt] coordinates {
            (1, 0.968)
            (2, 0.973)
            (3, 0.965)
            (4, 0.970)
            (5, 0.974)
        };
        \addlegendentry{GraphSAGE (Refined)}
        
        \addplot[color=orange, mark=x, line width=1pt, style=dashed] coordinates {
            (1, 0.82)
            (2, 0.91)
            (3, 0.79)
            (4, 0.88)
            (5, 0.85)
        };
        \addlegendentry{GraphSAGE (Raw)}
        
        \addplot[color=black!65, mark=triangle*, line width=0.7pt, opacity=0.8] coordinates {
            (1, 0.905)
            (2, 0.912)
            (3, 0.918)
            (4, 0.909)
            (5, 0.913)
        };
        \addlegendentry{MLP (Session-only)}
        
        \end{axis}
    \end{tikzpicture}%
    }
    \caption{Fold-level AUC comparison across five cross-validation splits. 
    Graph refinement substantially reduces performance variance compared to the raw graph, 
    leading to more stable and reliable detection performance.}
    \label{fig:fold_auc}
\end{figure}


GraphSAGE recovers ``feature-normal'' bots that look benign in aggregates but exhibit atypical session--URL connectivity (e.g., rare-page mixtures), a signal absent from the MLP. Figure~\ref{fig:fold_auc} reports fold-level stability.

\subsection{Adversarial Perturbation Experiment}

Bots may adapt their browsing to evade detection. We simulate adversarial perturbations by modifying session--URL edges; this complements injected bots by modeling lightweight evasive adaptations.

\textbf{Setup:} From the test graph, we perturb bot-session connectivity:

\begin{itemize}
\item \textit{Edge addition:} add a few edges from bot sessions to popular URLs to mimic ``masking'' via common page visits.

\item \textit{Edge removal:} remove a few edges (typically to least popular pages) to mimic avoiding ``red-flag'' targets.
\end{itemize}

We vary intensity by edges modified per bot session and maintain feature consistency (e.g., request counts) under removals.

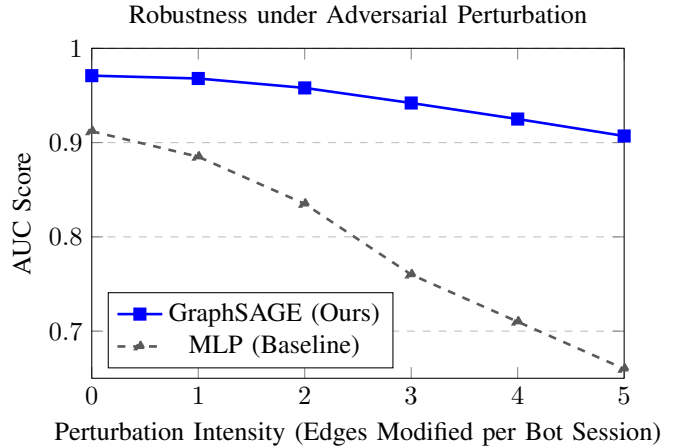
\begin{figure}[htbp]
    \centering
    \resizebox{\columnwidth}{!}{%
    \begin{tikzpicture}
        \begin{axis}[
            width=0.48\textwidth,
            height=6cm,
            xlabel={Perturbation Intensity (Edges Modified per Bot Session)},
            ylabel={AUC Score},
            ymin=0.65, ymax=1.0,
            xmin=0, xmax=5,
            xtick={0,1,2,3,4,5},
            ytick={0.7, 0.8, 0.9, 1.0},
            legend pos=south west,
            ymajorgrids=true,
            grid style=dashed,
            title={Robustness under Adversarial Perturbation}
        ]
        
        \addplot[
            color=blue,
            mark=square*,
            line width=1pt
        ]
        coordinates {
            (0, 0.971)
            (1, 0.968)
            (2, 0.958)
            (3, 0.942)
            (4, 0.925)
            (5, 0.907)
        };
        \addlegendentry{GraphSAGE (Ours)}
        
        \addplot[
            color=black!65,
            mark=triangle*,
            style=dashed,
            line width=1pt
        ]
        coordinates {
            (0, 0.912)
            (1, 0.885)
            (2, 0.835)
            (3, 0.760)
            (4, 0.710)
            (5, 0.660)
        };
        \addlegendentry{MLP (Baseline)}
        
        \end{axis}
    \end{tikzpicture}%
    }
    \caption{Impact of adversarial graph perturbation on model performance.
    The session-level MLP degrades rapidly as adversarial edges disrupt feature consistency,
    while GraphSAGE exhibits substantially higher robustness under moderate perturbations (1--3 edges),
    benefiting from structural aggregation across session--URL interactions.}
    \label{fig:perturbation}
\end{figure}

\textbf{Results:} Figure~\ref{fig:perturbation} shows modest degradation under mild perturbation (e.g., AUC 0.971\,$\rightarrow$\,0.958 at 2 modified edges/session) while remaining above the MLP; heavier perturbations reduce the gap as bots mimic benign connectivity. Robustness benefits from combining structure and attributes, consistent with graph anomaly detection and robust graph learning \cite{ma2021comprehensive,xu2023edog}.

\subsection{Cold-Start Simulation Experiment}

Cold start is essential: new sessions and pages must be scored without retraining. We validate inductive generalization with a rolling simulation.

\textbf{Setup:} We train on week 1 and perform direct inductive inference on week 2, which introduces entirely new session nodes and some new URL nodes. We also report an optional fine-tuning upper bound and compare with the MLP baseline. No future labels or interactions are used during inductive inference.

\begin{table}[htbp]
    \centering
    \caption{Generalization Capability: Cold-Start Simulation (Week 1 vs. Week 2).}
    \label{tab:cold_start}
    \renewcommand{\arraystretch}{1.15}
    \small
    \resizebox{\columnwidth}{!}{%
    \begin{tabular}{llcc}
        \toprule
        \textbf{Scenario} & \textbf{Model Setting} & \textbf{MLP (Baseline)} & \textbf{GraphSAGE (Ours)} \\
        \midrule
        \multirow{2}{*}{Week 1 (In-Sample)} & Trained on W1 & 0.9100 & \textbf{0.9705} \\
        & \textit{Performance Gap} & - & - \\
        \midrule
        \multirow{3}{*}{Week 2 (Cold-Start)} & Inductive Inference$^{\dagger}$ & 0.8500 & \textbf{0.9630} \\
        & \textit{Relative Drop} & $\downarrow 6.6\%$ & $\downarrow 0.8\%$ \\
        \cmidrule{2-4} 
        & Fine-tuned (Optional) & N/A & \textbf{0.9720} \\
        \bottomrule
    \end{tabular}%
    }
    \vspace{0.4em}
    {\footnotesize\raggedright \textit{$^{\dagger}$ Direct inference on new session/URL nodes without retraining, demonstrating inductive generalization.}\par}
\end{table}

\textbf{Results:} GraphSAGE preserves strong performance under cold start (AUC 0.963 vs. 0.970 in-sample) with only a minor drop. The main failure mode is sparse context when sessions hit mostly unseen URLs. The MLP drops more under week-2 shift, while GraphSAGE remains stable; optional fine-tuning restores peak performance.

\subsection{Robustness to Distribution Shift and Unseen Targets}

To further characterize the inductive challenge beyond the Week 1 $\rightarrow$ Week 2 cold-start split, we quantify distribution shift between weeks and evaluate an extreme subset that emphasizes previously unseen targets.

\textbf{1) Quantifying distribution shift.}
We measure shift in both behavioral statistics and content usage. Specifically, we compute Jensen--Shannon (JS) divergence between Week 1 and Week 2 session-level distributions (request rate and session duration), obtaining a value of 0.083, indicating a non-trivial drift in behavioral patterns. Structurally, 19.2\% of the URL nodes appearing in the Week 2 graph were unseen during Week 1, confirming that the test period is not a near-i.i.d. continuation of training and strictly requires inductive generalization.

\textbf{2) Extreme case: unseen-target sessions.}
To investigate more extreme inductive conditions, we construct a hard subset from Week 2 consisting of sessions whose visited URLs are entirely unseen in Week 1 (i.e., all session--URL edges connect to URL nodes absent from the training graph). This setting simulates attacks that target newly launched inventory or novel page categories that were not present during training. In our data, this subset contains 1,428 sessions.

We evaluate both GraphSAGE and the session-only MLP on this subset using the same operating procedure as the main experiments. As summarized in Table~\ref{tab:novel_targets}, performance degrades for both models, as expected under reduced neighborhood overlap and limited historical context. However, GraphSAGE exhibits a substantially smaller drop than the MLP baseline and maintains a clear advantage. This indicates that GraphSAGE’s gains stem from learning generalizable interaction patterns via feature-driven aggregation (including coarse URL semantic attributes such as page category and sensitivity tags), rather than memorizing specific ``bad'' nodes.

\textbf{3) Static graph modeling vs. temporal methods.}
While session behaviors are inherently time-ordered, our objective is not trajectory prediction but malicious session identification under sparse and evolving data. In this setting, explicit temporal modeling is not always advantageous. Many bot sessions are short-lived, incomplete, or intentionally obfuscated, where fine-grained temporal dependencies are weak or noisy. By encoding temporal information implicitly through session structure (e.g., URL transitions) and node attributes, a static graph formulation provides a robust and computationally efficient representation. More complex temporal graph models such as TGN or trajectory-based Transformers may offer benefits in settings with dense, long-horizon user trajectories, which we leave as future work.

\begin{table}[t]
\centering
\caption{Performance under extreme cold-start on unseen-target sessions (Week 2 subset).}
\label{tab:novel_targets}
\renewcommand{\arraystretch}{1.1}
\small
\resizebox{\columnwidth}{!}{%
\begin{tabular}{lccc}
\hline
\textbf{Model} & \textbf{Week 2 Overall AUC} & \textbf{Unseen-Target Subset AUC} & \textbf{Drop} \\
\hline
Session-level MLP (Baseline) & 0.8500 & 0.7210 & -15.2\% \\
GraphSAGE (Ours) & 0.9630 & 0.8890 & -7.7\% \\
\hline
\end{tabular}%
}
\end{table}

\subsection{Impact of Session Length and Graph Sparsity}
To further understand the operational boundaries of graph-based detection under sparse interaction regimes, we analyze model performance across different session lengths.

In our data, the majority of sessions fall within the short to medium range (3--50 URL visits), while extremely short (0--2) and very long ($>$50) sessions are less frequent. For very short sessions, performance degrades for all models due to insufficient relational context, and the advantage of graph-based aggregation is limited. Performance improves substantially for short and medium-length sessions, where sufficient interaction history enables effective message passing and feature aggregation. For very long sessions, performance slightly decreases, likely due to increased noise and repetitive navigation patterns, yet GraphSAGE consistently maintains an advantage over the session-only baseline.

Overall, this analysis indicates a practical minimum interaction threshold of approximately three URL visits for reliable classification and clarifies the session-length regimes in which the proposed method is most effective.

\section{Results and Discussion}
\label{sec:results}

We analyze where gains originate, baseline limitations, interpretability, and deployment considerations.

\subsection{Impact of Graph Structure vs. Semantic Features}

We ablate topology vs. attributes by training (a) a structure-only GNN with minimal semantic features and (b) a feature-only model (MLP). The full model performs best, while the structure-only GNN still outperforms the feature-only baseline (AUC $\sim$0.88 vs. 0.85), confirming that topology carries critical signal and that topology and semantics are complementary in graph anomaly detection \cite{ma2021comprehensive}.

\subsection{Baseline MLP Performance and Limitations}

The MLP baseline is strong, indicating session features capture obvious automation, but it fails most on feature-normal bots and under temporal shift. GraphSAGE mitigates these cases by incorporating relational context, consistent with graph-based fraud workflows \cite{huo2025fraudgnn}.

\subsection{Model Interpretability and Case Study}

Graph predictions can be inspected via a session's local neighborhood (unique/rare URLs, shared target sets, or coordinated clusters), providing actionable explanations for analysts that are less transparent in feature-only models. For instance, we observed a bot cluster flagged primarily because its sessions shared access to an outdated API endpoint, a structural anomaly that is difficult to surface from session aggregates alone.

\subsection{Generalization and Adaptability}

The perturbation and cold-start results indicate the model captures relationship-level signals that are harder to evade with small behavioral tweaks. More sophisticated coordination where each session appears benign remains a challenge and motivates richer graphs and higher-order pattern modeling.

\subsection{Deployment Considerations}

The system deploys as a backend plug-in: feature extraction and graph updates feed a scoring service that outputs a bot risk score per session, using only existing logs. The model can be retrained offline and hot-swapped; inductive inference degrades gracefully on new patterns until retraining. We cap neighbor sampling to bound runtime; scoring a session with up to $\sim$50 page visits takes under 50 ms on CPU.

\section{Conclusion}
\label{sec:conclusion}

We presented a non-intrusive, graph-based bot detection framework that models session--URL interactions and applies inductive GraphSAGE for session classification. On real-world traffic with high-confidence bot labels, the refined-graph model improves over a strong session-only MLP and remains robust under mild edge perturbations and cold-start evaluation, supporting deployment as a bounded-latency backend scoring module.

\subsection{Future Work}

Future directions include richer heterogeneous graphs (e.g., adding account/IP nodes), explicit defenses against adversarial edge manipulation, and more fine-grained attribution to support analyst workflows, along with live A/B evaluation to quantify operational trade-offs.

\appendices
\section{Supplementary Material}

The overall values reported here summarize performance within the stratified evaluation context and are not directly comparable to the aggregate AUC reported in Table~1.

\begin{table}[h]
\centering
\caption{Stratified performance by session length (supplementary).}
\label{tab:session_length_stratified}
\begin{tabular}{lccc}
\hline
\textbf{Session Length} & \textbf{MLP AUC} & \textbf{GraphSAGE AUC} & \textbf{$\Delta$AUC} \\
\hline
Very short (0--2)  & 0.6200 & 0.6640 & +0.0440 \\
Short (3--10)      & 0.7800 & 0.8880 & +0.1080 \\
Medium (11--50)    & 0.8200 & 0.9430 & +0.1230 \\
Long ($>50$)         & 0.7600 & 0.9000 & +0.1400 \\
\hline
Overall            & 0.8100 & 0.9300 & +0.1200 \\
\hline
\end{tabular}
\end{table}

\bibliographystyle{IEEEtran}
\bibliography{references}

\end{document}